\documentclass{article}

\usepackage[accepted]{icml2019}
\usepackage{times}

\usepackage{hyperref}       % hyperlinks

\usepackage{url}            % simple URL typesetting
\usepackage{booktabs}       % professional-quality tables
\usepackage{microtype}      % microtypography
\usepackage{amsmath}
\usepackage{graphicx}

\usepackage{xcolor}

\usepackage{acronym}
\newacro{tts}[TTS]{text-to-speech}
\newacro{mos}[MOS]{mean opinion score}

\usepackage{enumitem}

\newcommand{\vect}[1]{\boldsymbol{#1}}

\icmltitlerunning{CHiVE: Varying Prosody in Speech Synthesis}

\begin{document}

%% \maketitle

\twocolumn[
\icmltitle{CHiVE: Varying Prosody in Speech Synthesis with a Linguistically Driven Dynamic Hierarchical Conditional Variational Network}

\begin{icmlauthorlist}
\icmlauthor{Vincent Wan}{goo}
\icmlauthor{Chun-an Chan}{goo}
\icmlauthor{Tom Kenter}{goo}
\icmlauthor{Jakub Vit}{wba}
\icmlauthor{Rob Clark}{goo}
\end{icmlauthorlist}

\icmlaffiliation{goo}{TTS Research, Google UK, London}
\icmlaffiliation{wba}{University of West Bohemia, work carried out whilst at Google}

\icmlcorrespondingauthor{Tom Kenter}{tomkenter@google.com}

\vskip 0.3in
]

\printAffiliationsAndNotice{}

\begin{abstract}
The prosodic aspects of speech signals produced by current text-to-speech systems are typically averaged over training material, and as such lack the variety and liveliness found in natural speech.
To avoid monotony and averaged prosody contours, it is desirable to have a way of modeling the variation in the prosodic  aspects of speech, so audio signals can be synthesized in multiple ways for a given text.
We present a new, hierarchically structured conditional variational auto-encoder to generate prosodic features (fundamental frequency, energy and duration) suitable for use with a vocoder or a generative model like WaveNet.
% The encoder compresses linguistic and acoustic input features into a variational bottleneck layer.
% The decoder has the same linguistic features as then input encoder, but, instead of having access to acoustic features directly, gets an embedding sampled from the variational layer, designed to capture the prosodic aspects of the input text.
At inference time, an embedding representing the prosody of a sentence may be sampled from the variational layer to allow for prosodic variation.
To efficiently capture the hierarchical nature of the linguistic input (words, syllables and phones), both the encoder and decoder parts of the auto-encoder are hierarchical, in line with the linguistic structure, with layers being clocked dynamically at the respective rates. 
We show in our experiments that our dynamic hierarchical network outperforms a non-hierarchical state-of-the-art baseline, and, additionally, that prosody transfer across sentences is possible by employing the prosody embedding of one sentence to generate the speech signal of another.
\end{abstract}

\section{Introduction} % (fold)
\label{sec:introduction}

Most current \ac{tts} prosody modeling paradigms implicitly assume a one-to-one mapping between text and prosody and fail to recognize the one-to-many nature of the task, i.e., there is a large number of ways in which a given sequence of words can be spoken.
This leads to an averaging effect and as such a lack of variation and variety in generated synthetic speech. This work aims to model prosody in a way that avoids this problem and works towards allowing mechanisms to control the variation in the generated prosody in a linguistically motivated way.

Prosody prediction in \ac{tts} is usually concerned with providing segment durations and fundamental frequency (\(F_0\)) contours for the utterance being synthesized. These targets are used to guide unit selection \cite{fujii2003target}, form the input features to a vocoder \cite{yoshimura1999simultaneous}, or drive a WaveNet-like model \cite{oord2016wavenet}.
The way in which prosody models are traditionally built has a number of inherent
problems.
Firstly, as stated above, there is an underlying assumption that there is a unique mapping from a given text or linguistic specification to a prosodic realization.
Secondly, duration is modeled independently of \(F_0\), e.g., \cite{ZenAEHS16}.
Lastly, modeling techniques are generally frame- or phone-based and as such, lack the ability to ensure linguistically valid prosodic contours for an utterance as a whole.

\begin{figure}[t]
\centering
\includegraphics[width=.48\textwidth]{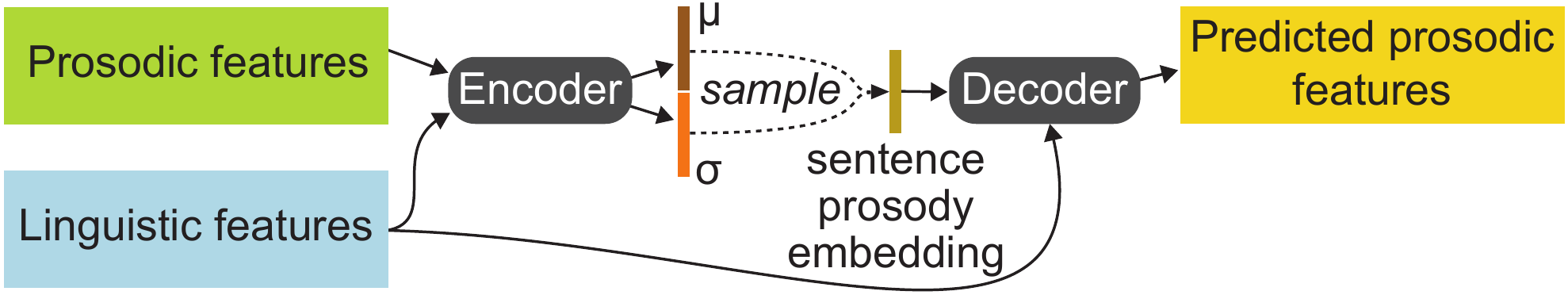}
\caption{High-level overview of the CHiVE model at training time.}
\label{fig:chive_high_level}
\end{figure}
In this paper we propose a new model to explicitly address the problems described above.
We present a model called a Clockwork Hierarchical Variational autoEncoder (CHiVE), that produces \(F_0\) and duration and, additionally, $c_0$ (the 0th cepstrum coefficient) as energy, which is a strong prosodic correlate.
In our work, these predicted prosodic features are used to drive a WaveNet model \cite{oord2016wavenet} to produce the final speech signal.

The CHiVE model is trained as a conditional variational auto-encoder \cite{kingma2014vae,NIPS2015_5775}.
A high-level graphical overview is provided in Figure~\ref{fig:chive_high_level}.
The encoder has both linguistic features (a matrix represented by a blue rectangle) and prosodic features (green rectangle) as input, and encodes these into an embedding using a variational bottleneck layer.
The linguistic features represent aspects of the input such as part-of-speech for words, syllable information and phone-level information.
The prosodic features represent acoustic information about the input, in terms of pitch, duration and energy.
The variational layer predicts two vectors, which are interpreted as means and variances (the brown and orange vectors in Figure~\ref{fig:chive_high_level} respectively), used to parameterize an isotropic Gaussian distribution, from which an embedding is sampled.
The decoder has linguistic features as input (the same ones the encoder has), and is additionally fed an embedding sampled from the variational layer.
The sampled embedding, in this way, is designed to encode prosodic aspects of the input sentence.
We refer to it as the \emph{sentence prosody embedding}.

To capture the linguistic hierarchy incorporated in the input features -- provided at the level of words, syllables, phones and frames -- both the encoder and the decoder are hierarchical, with dynamically clocked layers following the layers in the linguistic structure of an utterance.
A unique feature of the CHiVE mdoel is that the structure of the network is dynamic, i.e., the unrolled recurrent structure of the model is different for every input sentence, depending on the linguistic structure of that sentence.
The model is described in full in \S\ref{sec:model_description}.
We show in our experiments that our dynamic hierarchical network outperforms a non-hierarchical baseline.

The main value of CHiVE lies in the distribution learned by the variational layer, which represents a space of valid prosodic specifications.
At inference time, two different scenarios can be employed to make use of this space.\newline
The first scenario is where we discard the encoder, and choose or sample a sentence prosody embedding from the Gaussian prior, trained at the variational layer. The zero vector can be chosen, which, by design, is the mean of the sentence prosody embedding space, and in our experience tends to represent prosody which exhibits the averaging effect as seen in models such as \cite{ZenAEHS16}.\newline
The second scenario is similar to the setup at training time, where a sentence is encoded by the encoder.
Yet, instead of trying to reproduce the prosodic features of the input sentence as in the auto-encoder setting during training, we can use the predicted sentence prosody embedding to generate speech for a different sentence, thereby transferring the prosody of one sentence to another. 
In this setting, the decoder gets the linguistic features of the new sentence as input, while it is conditioned on a sentence prosody embedding of a reference sentence.
This way, the speech for the textual content of the new sentence will be synthesized with the prosody of the reference sentence.
% Lastly, there are internal representations within the CHiVE model that represent the prosody at other linguistic levels, e.g., syllables.
% These might make it possible to directly influence, or overwrite, prosody at these levels, although doing so is beyond the scope of this paper.

In short, our contributions are:
\setlist{nolistsep}
\begin{itemize}
  \itemsep=.4em
\item We present CHiVE, a linguistically driven dynamic hierarchical conditional variational auto-encoder, the first of its kind to have a dynamic network layout that depends on the linguistic structure of its input.  
  \item We show that the CHiVE model yields a meaningful prosodic space, by showing that prosodic features sampled from it yield synthesized audio superior to the audio based on features sampled from a non-hierarchical variational baseline.
\item We illustrate prosody transfer, where the prosody of a reference sentence, as captured by the variational layer in the CHiVE model, can be used to synthesize speech for another sentence.
\end{itemize}

\section{Related Work} % (fold)
\label{sec:related_work}

We discuss work related to three aspects of the CHiVE model we present: modelling of prosody, hierarchcial models and variational models. 

\subsection{Modeling prosody}

%% TK: Skipped some of this and moved it down.
%% For historical reasons, \ac{tts} systems generally model duration independently of other aspects of prosody.
%% Moreover, as prosodic aspects were typically modeled at the level of phones or syllables, without taking the entire sequence into account, long term prosodic effects were neglected.
%% In contrast to the early work described below, our approach models the primary acoustic prosodic features used to realize prosody jointly, and across the full sequence.

Early parametric approaches to \ac{tts} using hidden Markov models (HMMs) adopted a ``bucket of Gaussians'' approach to duration modeling~\cite{yoshimura1998duration} which easily fits within the HMM architecture. This could be considered a step backwards from earlier approaches, for example:  \cite{black1998festival, van1994assignment, goubanova2008bayesian}  Additionaly, in these models $F_0$ is considered as just another acoustic feature, i.e., a local feature of a context-dependent phone, rather than a supra-segmental property of the utterance;  earlier work including \cite{fujisaki1984analysis, wblagenerating,vainio1999neural} was ignored.\newline
This trend continued with the move to deep neural networks \cite{ze2013statistical, zen2015unidirectional}.
There are a few exceptions.
In \cite{mixdorff2001building} a feed forward neural network is presented to predict prosody using a Fujisaki-style model, but they only apply it to re-synthesis rather than \ac{tts};
In \cite{tokuda2016temporal}, an integrated model is proposed using a hidden semi-Markov model within a neural network framework. 
In contrast to the early work described above work, our approach models the primary acoustic prosodic features used to realize prosody jointly, and across the full sequence.

An alternative approach to modeling prosody explicitly is taken by end-to-end approaches to \ac{tts}, for example, \cite{deepvoice3, wang2017tacotron} where there is no explicit prosody model at all, and any modeling of prosody is performed implicitly within the model.
To a large extent these end-to-end based approaches appear to be able to generate natural prosody and expression, however, in their current forms it is not clear how to obtain varied prosodic patterns.
% or control the prosody in any way. % TK: We don't do that either (in this paper).
Recent work with style tokens~\cite{wang2018style_tokens} attempts to address this point, but the style that can be controlled so far is quite abstract.
Style tokens may guide utterance level prosodic features such as emotion, pitch range, or speaking-rate, but the representations learned by the tokens do not necessarily represent clear useful styles.
In short, the task style tokens are designed for -- capturing a particular style across many utterances -- is different from the one we address here: varying the intonational aspects of prosody per utterance.
% It is not immediately obvious how the style token work should be adapted so it can be applied to our task.

\subsection{Hierarchical models}

Hierarchical models are popular in natural language processing, e.g., to encode sentence embeddings from the sequence of words in them \cite{serban2017hierarchical,amn2017kenter}, or to construct sentence representation from a sequence of words, which are in turn treated as sequences of characters, or bytes \cite{jozefowicz2016exploring,byte-level2018kenter}.
In \ac{tts}, hierarchical networks have been explored to a limited extent:
\cite{srikanth2017hierarchical} use hierarchical information by having recurrent connections at frame and phone levels.
In other work, \cite{Ribeiro2016parallelcascaded, YIN201682} try to integrate hierarchical information, but in the context of feed-forward neural networks instead of recurrent neural networks (RNNs).\newline
Most similar to our approach are the clockwork models such as \cite{koutnik2014clockwork}, from which we derive the name.
Different from that work, the clockwork model we present is run as a conditional variational auto-encoder, rather than as a standard encoder-decoder model.
Lastly, \cite{liu2015multi} have a set of fixed clocking speeds that are independent of the input sequence and act within sub-portions of a layer of RNN cells.
In contrast to that work, we adopt dynamic clock rates across layers by utilizing sequence-to-sequence encoding techniques.

\subsection{Variational models}

There are few recent attempts to use VAE-style models in speech synthesis. For example \cite{akuzawa2018expressive} propose VAE-Loop where the latent space is used to model expression of speech, and \cite{henter2018deep} use a number of VAE related techniques to show that categories of acted emotion can be modeled in an unsupervised fashion. 
\cite{hsu2018hierarchical} present an extended Tacotron approach combining a variational approach with a basic hierarchical structure to better address separation of latent feature spaces.
Again these techniques are concerned with the abstract nature of the utterance prosody such as speaking style rather than being concerned with the details of prosodic meaning. 

\section{The CHiVE model}
\label{sec:model_description}

The CHiVE model learns a mapping between pairs of \(\{\vect{x}, \vect{y}\}\), where $\vect{x}$ is a sequence of input features, and $\vect{y}$ a sequence of prosodic parameters.
The features describing the linguistic aspects of the input are denoted as $\vect{x}^{linguistic}$.
They are provided in a hierarchical fashion -- a tree structure -- with features at sentence level, word level, syllable level, phone level and frame level.
We denote the features pertaining to the acoustic prosodic information ($F_0$, $c_0$ and duration) as $\vect{x}^{prosodic}$, consisting of $\vect{x}^{F_0c_0}$ and $\vect{x}^{duration}$.
The prosodic features are presented at multiple levels as well.
Throughout this paper, let \(p \in [0, \ldots, P]\) enumerate phones, and \(t \in [0, \ldots, T]\) enumerate  acoustic frames in sequences \(\vect{x}\) and $\vect{y}$.
Duration values, $\vect{x}^{duration}_p$, are provided at phone level, while $F_0$ and $c_0$ values, $\vect{x}^{F_0c_0}_t$, are provided at frame level.
We note that $\vect{y} = \vect{x}^{prosodic}$, i.e., as the CHiVE model is a (conditional) auto-encoder, its task is to reconstruct the input features $\vect{x}^{prosodic}$.

Predicting the two output sequences, $\vect{\hat{y}}^{duration}$ and $\vect{\hat{y}}^{F_0c_0}$, is modeled as a regression task.
The predictions are computed from the input features by two hierarchical recurrent neural networks (RNNs), with a variational layer in between:
\begin{eqnarray*}
  \vect{\hat{\mu}}, \vect{\hat{\sigma}} & = & \mathcal{H}^{encoder}(\vect{x}^{prosodic}, \vect{x}^{linguistic}) \\
  \vect{\hat{s}} & \sim & \mathcal{N}(\vect{\hat{\mu}}, \vect{\hat{\sigma}}) \\
  \vect{\hat{y}}^{duration}, \vect{\hat{y}}^{F_0c_0} & = & \mathcal{H}^{decoder}(\vect{\hat{s}}, \vect{x}^{linguistic}).
\end{eqnarray*}
The hierarchical encoder and decoder, and the way they handle features, are described in more detail in \S\ref{sec:encoder} and \S\ref{sec:decoder}, respectively.
The sentence prosody embedding $\vect{\hat{s}}$ is sampled from an isotropic multi-dimensional Gaussian distribution, parametrized by predicted mean vector $\vect{\hat{\mu}}$ and variance vector $\vect{\hat{\sigma}}$.

The network is optimized to minimize two \(L_2\) losses, plus a KL divergence term for the variational layer:
\begin{eqnarray}\label{eqn:model_loss}
  \mathcal{L} & = & ~~~~\lambda_1 \sum_p||\vect{y}^{duration}_p - \vect{\hat{y}}^{duration}_p||^2 \\
  & & +~\lambda_2 \sum_t||\vect{y}^{F_0c_0}_t - \vect{\hat{y}}^{F_0c_0}_t||^2 \nonumber \\
  & & +~\lambda_3 D_{\textit{KL}}[\mathcal{N}(\vect{\hat{\mu}}, \vect{\hat{\sigma}})~||~\mathcal{N}(0, 1)], \nonumber
\end{eqnarray}
where the first $L_2$ loss is computed over the durations values per phone, the second \(L_2\) loss is computed over the \(F_0\) and \(c_0\) values per frame, and $\vect{\hat{\mu}}$ and $\vect{\hat{\sigma}}$ are the means and variances as predicted by the encoder, respectively.
The $\lambda$ terms are used to weight the three losses.

% A common problem when training variational layers is that the KL-divergence term makes it hard for the encoder to start learning anything initially \todo{REFS}.
% This is remedied in the usual way here, by introducing a $\lambda$ term that allows for the KL-divergence term to be weighted differently across time steps \todo{REFS}.
% In particular, for training step $i$ and a minimum $m$ and maximum value $M$, we have:
% %
% \begin{equation}\label{eqn:kl_weight}
%   \lambda = \begin{cases}
%     0, & \text{if}~i < m, \\
%     \frac{i}{M}, & \text{if}~m < i < M, \\
%     1, & \text{if}~i > M.
%     \end{cases}
% \end{equation}
%
The CHiVE model consists of three parts: an encoder, a variational layer, and a decoder.
We will describe the design of each of these in turn in this section.
Additional details about the hyper-parameters of the network and the features used are discussed in \S\ref{sec:experimental_setup}.

\subsection{CHiVE encoder}
\label{sec:encoder}

\begin{figure*}[ht]
\centering
\includegraphics[width=.9\textwidth]{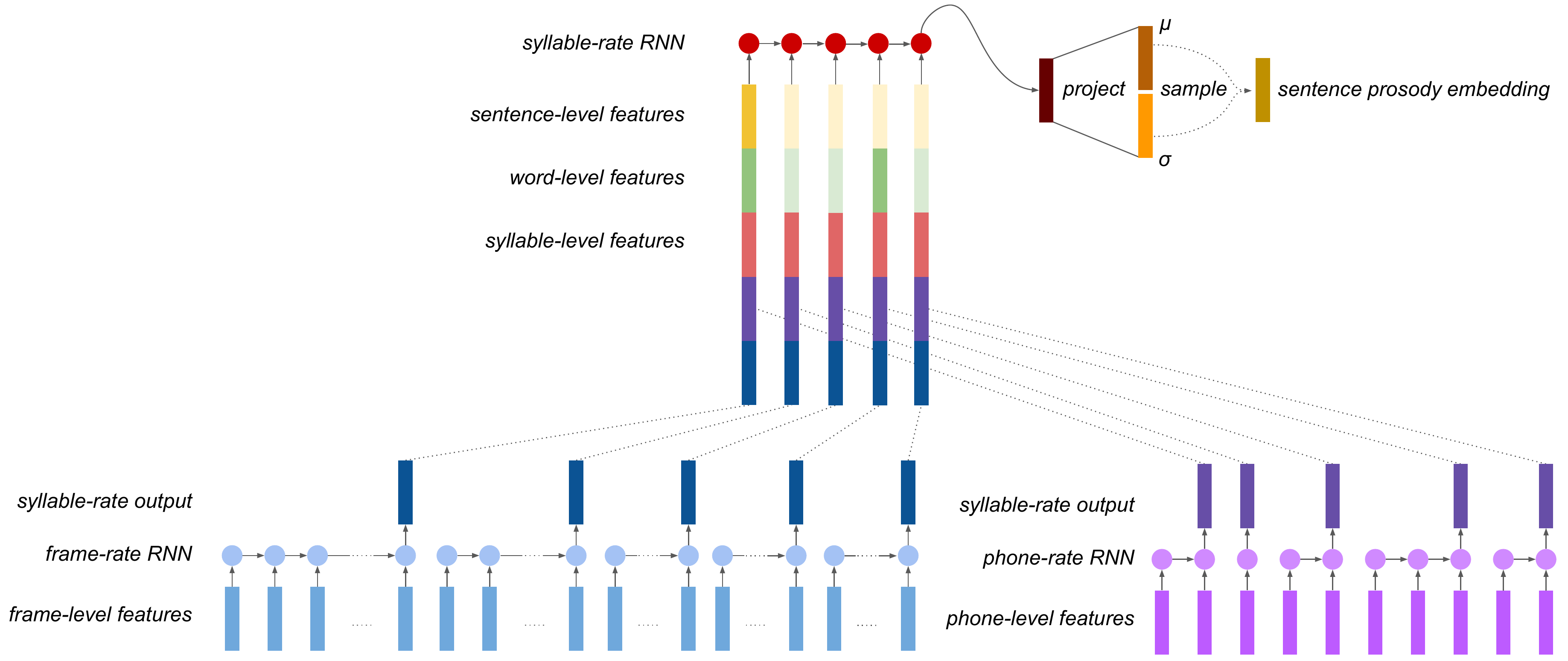}
\caption{CHiVE encoder and variational layer. Circles represent blocks of RNN cells, rectangles represent vectors. Broadcasting (vector duplication) is indicated by displaying vectors in a lighter shade of the same colour. (Best viewed in colour).}
\label{fig:chive_encoder}
\end{figure*}
The CHiVE encoder is a hierarchical RNN, composed of three RNNs: a frame-rate RNN and a phone-rate RNN, that both produce syllable-rate output, and a syllable-rate RNN that takes in the output of the other two RNNs, in addition to sentence-, word- and syllable-level linguistic input features.
Figure~\ref{fig:chive_encoder} gives a graphical depiction of the CHiVE encoder.

The frame-rate RNN, depicted in blue, reads frame-level prosodic features (\(F_0\) and \(c_0\)), and outputs its internal activation at the end of every syllable. There are usually many frames in a syllable -- our experiments use a 5 millisecond frame shift -- most of which are left out of the figure for clarity.  The phone-rate RNN runs asynchronously in parallel, and like the frame-rate RNN emits its hidden state at every syllable boundary. In the figure the first syllable consists of two phones and the second syllable of one phone. The states of both LSTMs are reset at syllable boundaries.\newline
The outputs of the two RNNs are concatenated and joined with additional syllable-, word- and sentence-level input features.
The word-level and sentence-level input features are broadcast as appropriate, which is indicated in Figure~\ref{fig:chive_encoder} by them having a slightly lighter shade.
As we can see from the figure, there is only one sentence-level feature vector, broadcast to every time step.
Similarly, there are two words, the feature vectors of which are broadcast across their respective syllables.
Finally, the last hidden activation of the syllable-rate RNN is taken to represent the entire input and passed on to a variational layer.

The hierarchical layout of the model is designed to reflect the hierarchical nature of the linguistic information that the features represent.
We should note two things concerning the clock rates of the RNNs.
Firstly, the number of input feature vectors for the RNN running at frame-rate is different from the number of input feature vectors for the one running at phone-rate.
Nonetheless both RNNs produce the same number of output vectors: one for every syllable.
Secondly, the number of input feature vectors at all levels (frames, phones, syllables and words) is different across sentences.
The CHiVE model handles this appropriately, by letting each sub-RNN run at the right clock speed, determined by the number of features provided at that level by the input structure.
As such, CHiVE dynamically follows the linguistic structure of the input sentence, which is the main feature setting it apart from other models.

\subsection{Variational layer}

The variational layer takes the last hidden activation of the syllable-rate RNN (depicted in dark red in Figure~\ref{fig:chive_encoder}) as input, passes it through a fully connected layer, and splits the resulting vector into two vectors, representing the mean and variance of an isotropic multi-variate Gaussian (displayed in brown and orange, respectively, in the figure).
Finally, a vector is sampled from this Gaussian (displayed in dark yellow), which is used as input to the decoder.

\subsection{CHiVE decoder}
\label{sec:decoder}

\begin{figure}[ht]
\centering
\includegraphics[width=0.45\textwidth]{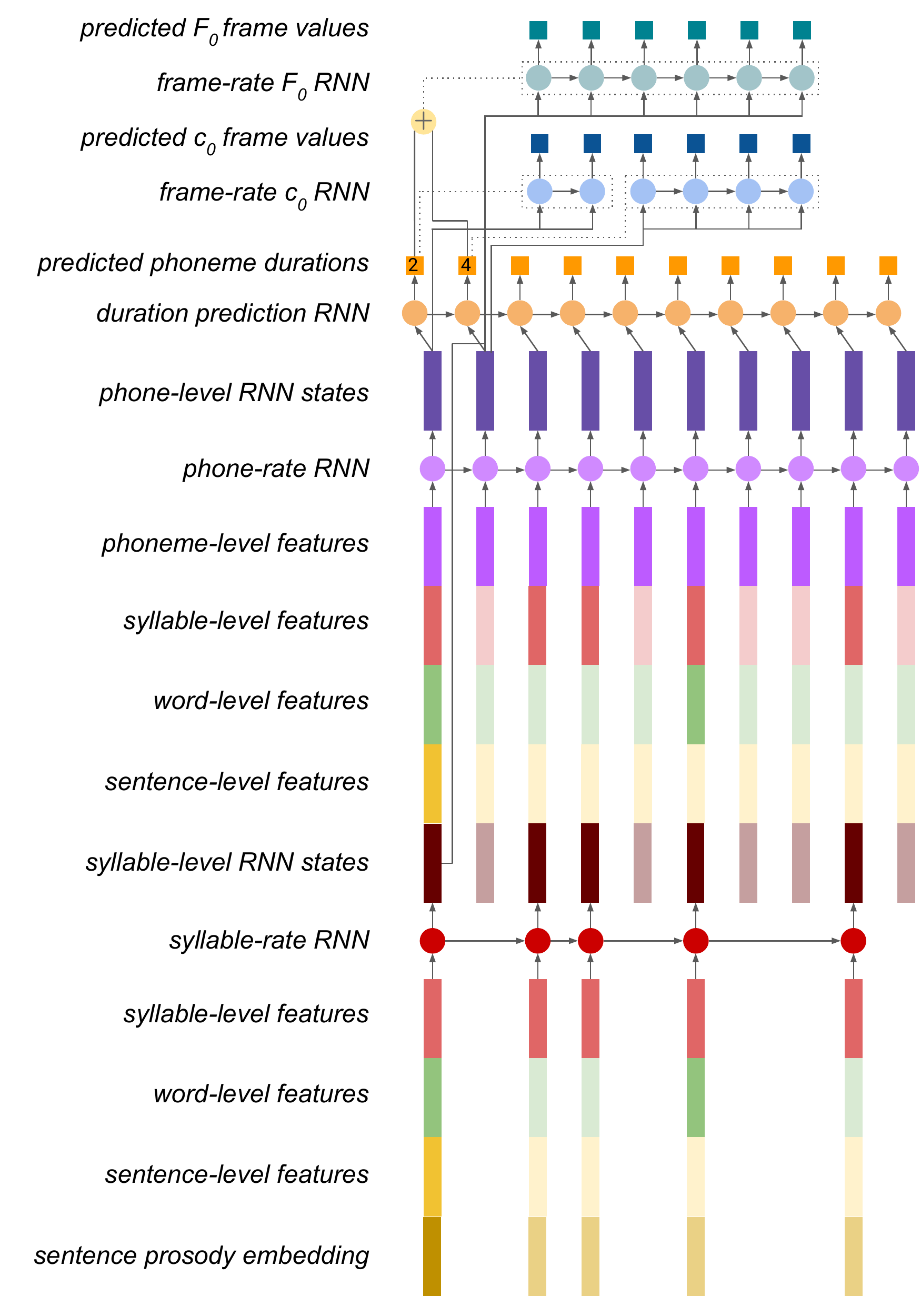}
\caption{CHiVE decoder. Circles represent blocks of RNN cells, rectangles represent vectors. Broadcasting (vector duplication) is indicated by displaying vectors in a lighter shade of the same colour. (Best viewed in colour).}
\label{fig:chive_decoder}
\end{figure}
The CHiVE decoder is similar to the encoder, but the hierarchy is reversed.
The decoder goes from sentence-level to the levels of syllables, phones and frames.
Figure~\ref{fig:chive_decoder} gives a graphical overview of the decoder.

The sentence prosody embedding output by the variational layer in the previous step (displayed in dark yellow again in the figure) is concatenated with sentence-level (yellow), word-level (green) and syllable-level features (pink), all broadcast as appropriate, and used as input to a syllable-rate RNN that produces one output per syllable.
To reiterate, as we are auto-encoding, the linguistic feature vectors here are the same as those input to the encoder.

The next level of the decoder is a phone-rate RNN (in purple), that takes as input each output of the syllable-rate RNN,  concatenated with sentence-level features, word-level features, syllable-level features (the same ones that fed into the syllable-level RNN, simply read again at this stage) and phone-level features to produce a sequence of phone representations. The output activations of the phone-rate RNN (in dark purple) are used as input to three different RNNs, that together model the prosodic features, $F_0$, $c_0$ and duration, as highlighted in \S\ref{sec:introduction}.\newline
Firstly, an RNN predicts phone duration (displayed in orange), expressed as the number of frames per phone.
In Figure~\ref{fig:chive_decoder} the predicted values for the first two phones are 2 and 4, respectively (these numbers are low to keep the figure clear; in  reality these numbers are much higher).
Secondly, to predict the \(c_0\) values for each frame, another RNN (in blue) is run for as many steps as predicted by the phone duration RNN, where the corresponding phone-rate RNN output activation is repeatedly fed as input at every time step. In the figure, the \(c_0\) RNNs for the first two phones are displayed, which unroll for 2 time steps and 4 time steps for the first and second phone, respectively.
The last state of one RNN is used as initial state for the next.
In reality, the \(c_0\) RNN is run for every phone but for reasons of clarity this is not shown in the figure.
Finally, the \(F_0\) values are predicted by another frame-rate RNN (in turquoise), for every frame in each syllable. To accomplish this, the durations for all phones in a syllable are summed, and an RNN predicting \(F_0\) values is unrolled for as many steps.
So, instead of unrolling this RNN twice -- once for 2 steps and once for 4 steps as is the case for the \(c_0\) prediction -- the RNN is run once for each syllable, the unroll length of which is the sum of the durations of its constituent phones -- 6, for the first syllable in our example in Figure~\ref{fig:chive_decoder}.
The input at every time step is the syllable-rate RNN activation (dark red), concatenated with the phone-rate RNN activation (dark purple) corresponding to the end of the syllable.
In Figure~\ref{fig:chive_decoder} this is the second phone, as the first syllable consists of two phones.
Again, the $F_0$ RNN does run for all other syllables too, yet to keep the figure clear, only one frame-level $F_0$ RNN is displayed.

With respect to the duration values, it should be noted that during training the \emph{predicted} durations are used for calculating the $L_2$ loss, but the \emph{ground truth} durations are used to determine the unroll lengths of the frame-level $c_0$ and $F_0$ RNNs.
This allows for easy calculation of the differences between predicted $c_0$ and $F_0$ values, which can be computed by a one-to-one matching of frames.
At inference time there is no ground truth duration available, and the predicted duration values are used.

In addition to the features described above, every RNN level is provided with a timing signal, using a standard cosine coarse encoding technique \cite{ze2013statistical}.
Different dimensions of coding signal are used for different hierarchy levels: 64 for words, 4 for syllables and phones, and 3 for frames.
These timing vectors can be thought of as being part of the features vectors at every level in the model, and are therefore not explicitly depicted in Figures~\ref{fig:chive_encoder}~and~\ref{fig:chive_decoder}.

\section{Experiments}
\label{sec:experiments}

Our main experiment compares the prosodic features sampled from the distribution yielded by our hierarchical variational model to those sampled from a non-hierarchical baseline model.
In addition, we illustrate that CHiVE can be used to transfer the prosody of one sentence to another.\newline
Speech samples for the main and additional experiments, and for the prosody transfer illustrations described below are available at {\small \url{https://google.github.io/chive-varying-prosody-icml-2019/}}.

\subsection{Sampling prosodic features}
\label{sec:main_experiment}

As described in \S\ref{sec:introduction}, the aim of CHiVE is to yield a prosodic space from which suitable features can be sampled, to be used by a generative model producing speech audio.
We compare our dynamic hierarchical variational network to a non-hierarchical variational network, where the features generated by both are used to drive the same generative model to produce speech audio. 

\subsubsection{Experimental methodology}
\label{sec:experimental_setup}

\textbf{Baseline}~~
A non-hierarchical baseline was constructed to mirror the CHiVE model but without the dynamic structure.
It was a single multi-layer model running at frame time steps, with broadcasting at different levels, but no dynamic encoding/decoding.

\textbf{Test set}~~
The test set consisted of 100 sentences with 10 prosodically different renditions having been synthesized for each, both for our hierarchical CHiVE model, and for the non-hierarchical baseline.
That is, for each sentence we sampled 10 sentence prosody embeddings from each model and run the associated decoder to produce prosodic features, which we then supplied to Parallel WaveNet~\cite{oord2017parallel} to produce speech.
The WaveNet model made use of the duration and $F_0$ features, but not $c_0$.
We then randomly paired audio samples of the same sentence, one from each model, which gave us 1000 test sample pairs. We performed a side-by-side AB comparison, where raters were asked to indicate which of two audio samples of a pair they thought was better. Each rater received a random selection of pairs to rate and each pair was rated once.
\begin{table}[t]
  \caption{Side-by-side comparison between CHiVE and a baseline non-hierarchical model. Results are found to be significant, with a normal approximation to a binomial test z-value=-5.5, and p-value=3.91$\times \text{10}^{\text{-8}}$.}
  \begin{tabular}{c c c}
    \toprule
    Baseline preferred & neutral & CHiVE preferred \\
    \midrule
    292 (30.7\%) & 220 (23.2\%) & \textbf{438} (46.1\%) \\
    \bottomrule
  \end{tabular}
  \label{table:main_results}
\end{table}

\textbf{Training and eval set}~~
Our training data was recorded by 22 American English speakers in studio conditions, the number of male and female speakers being balanced.
The data consists of 161\,000 lines, from a range of domains, including jokes, poems, Wikipedia and other web data.\newline
The evaluation data for hyperparameter tuning was a set  of 100 randomly selected sentences per speaker, held out from training.
The input features used include information about phoneme identity, number of syllables, dependency parse tree.
We closely followed previous work \cite{zen2015unidirectional}, with two main differences: 1) a one-hot speaker identification feature was added and 2) features are split into levels corresponding to the levels of the linguistic structure.

\textbf{Hyperparameters}~~
Both the CHiVE prosody model and the non-hierarchical baseline use LSTM cells.
The RNNs at each level of the hierarchy consist of two LSTM layers.
The variational layer predicts the parameters of a 256-dimensional Gaussian distribution.
The hidden layers of the syllable and phone-level RNNs have dimension 32 at every layer, both in case of the the encoder and the decoder. We use a batch size of 4 for training.

\subsubsection{Results}

Table~\ref{table:main_results} lists the results of the side-by-side comparison between our CHiVE model and the non-hierarchical baseline model.
The speech generated using the CHiVE prosody model is significantly preferred over the speech generated using the baseline prosody model. We conclude that incorporating the hierarchical structure of the linguistic input, as done in CHiVE, yields better performance than processing the exact same features in a non-hierarchical fashion. 

\subsection{Further Evaluation}

To further evaluate CHiVE's performance we perform \ac{mos} tests.
Additionally, to gain insight into the training phase, we analyse the models in terms of loss.

\textbf{MOS listening tests}~~
\begin{table}[t]
  \caption{Results of \ac{mos} test comparing CHiVE to the baseline model. The interval shown is 95\% confidence around the mean.}
  \centering
  \begin{tabular}{l c}
    \toprule
     & MOS \\
    \midrule
    Baseline based on \cite{ZenAEHS16} & 4.01 \(\pm\) 0.11 \\
    CHiVE - zero embedding             & 4.07 \(\pm\) 0.10 \\
    CHiVE - encoded embedding          & 4.25 \(\pm\) 0.10 \\
    \midrule
    Real speech & 4.67 \(\pm\) 0.07 \\
    \bottomrule
  \end{tabular}
  \label{tab:mos_results}
\end{table}
To evaluate the quality of the different methods of obtaining sentence level embeddings we performed a series of \ac{mos} listening tests.
We compare using an all-zero sentence prosody embedding to an embedding made by encoding the ground truth audio. As a top-line we include the ground truth audio itself, and as a baseline we include the model from \cite{ZenAEHS16} as used by WaveNet \cite{oord2017parallel} to represent current state of the art. We choose a test set of 200 expressive sentences including jokes, poems and trivia game prompts as prosody is an important aspect of their rendition and we want to see how well our model is able to capture this. Raters judge each sentence for naturalness.

Results are shown in Table~\ref{tab:mos_results} where we see that the zero embedding is judged to be slightly, but not significantly, better than the baseline.
The encoded version is better still, which is unsurprising as it is using the auto-encoder properties of the model. 
The performance here is still lower than that of the real speech.  This might be caused by  the acoustic information being compressed in the variational layer, the fact that a sentence prosody embedding is sampled from the encoded distribution or highly expressive properties, such as increased energy or vocal fry, are still lost by the waveform generation model as it is currently configured to make use only of the predicted \(F_0\) and duration values. We conclude that while using the zero vector to drive prosody generation yields reasonable natural intonation, it may lack the true expressiveness that the model is capable of generating. We investigate this further below.  

\begin{figure}[t]
  \centering
  \includegraphics[width=.475\textwidth]{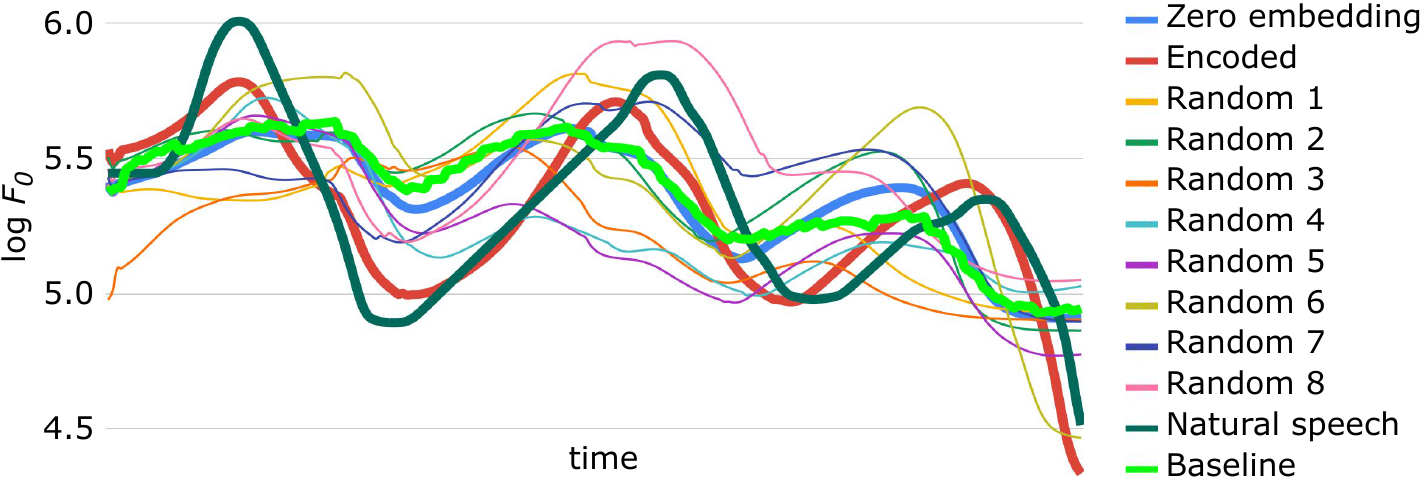}
  \caption{A selection of randomly generated \(\log F_0\) contours produced by CHiVE for the sentence ``That's a super choice!''. Contours do not align perfectly on the time axis because the predicted phone durations differ per contour. (Best viewed in colour).}
  \label{fig:hol_variation}
\end{figure}

\textbf{Variation of a single utterance}~~
Figure~\ref{fig:hol_variation} shows a range of sampled and reference \(\log F_0\) contours for a single sentence.
We first note that the natural speech contour and contour generated by CHiVE in VAE mode, labeled 'Encoded', are both very similar and expressive. This shows that the model works well as an encoder-decoder.  Secondly we note that the zero-embedding and baseline contours are very similar, and less expressive than the natural speech and (auto)encoded speech. This demonstrates the averaging effect of the same text spoken in different ways---the peaks representing pitch events are broader and less distinct.
Finally we see that the randomly sampled sentence prosody embeddings in general produce more expressive speech, and produce a wide range of different prosodic patterns, with prominence in different places. While we have not specifically evaluated the naturalness of these examples, anecdotal evidence suggests none of them sound unnatural.

\textbf{Performance during training}~~
We present RMSE and absolute error for our hierarchical CHiVE model and the non-hierarchical baseline used in our main experiment, described in section~\ref{sec:experimental_setup}.
The results presented are calculated on a held-out evaluation set.
As noted earlier -- when discussing Equation~\ref{eqn:model_loss}, \S\ref{sec:model_description} -- the $L_2$ losses are computed over $F_0$, $c_0$ and duration.
A breakdown of the overall loss is shown in Table~\ref{tab:objective_results}. 
For the line marked 'enc.', the sampling step is skipped and the mean values, $\vect{\mu}$, predicted by the encoder are directly given as input to the decoder.
That is, we specifically evaluate both models' ability to predict suitable durations and \(F_0\) contour given the full information of the held-out example being evaluated.
%% tomkenter: I am not sure what this line is supposed to add...
%% Note there is still scope for variation of the prosody for a given sentence as it is modeled as a distribution from which we sample.

From Table~\ref{tab:objective_results}, we see a 21\% relative reduction in the \(\log F_0\) RMSE compared to the baseline when the mean values predicted by the encoder are used.
\begin{table}[!t]
  \caption{Loss terms of the non-hierarchical baseline used in main experiment (lines labelled Bsl) and CHiVE, when an embedding sampled from the encoder is used (enc.), a zero embedding (0s), or a random embedding (rnd.).
    Both absolute error (Abs.) and root mean squared error (RMSE) are shown to allow comparison to other work.}
  \centering
  \begin{tabular}{l c c c c}
    \toprule
    & \(\log F_0\)      & \(F_0\)                 & $c_0$              & Duration \\
    & {\scriptsize RMSE} & {\scriptsize Abs.\thinspace(Hz)} & {\scriptsize RMSE} & {\scriptsize RMSE/Abs.\thinspace(ms)} \\
    \midrule
    % https://mldash.corp.google.com/compare?eidstrs=5519975120968818066,6495314477129930718&tag=Sentence_embedding_F0_mean_abs_error_Hz_all_segments&runsRegex=eval&numToShow=10
    %
    % Results at 1M iterations, as those were used for the main experiment too.
    %
    % Log_F0_RMSE_all_segments_
    % Sentence_embedding_F0_mean_abs_error_Hz_all_segments
    % Unscaled_C0_RMSE
    % Unscaled_duration_RMSE_from_sentence_embeddings/Duration_RMSE_seconds 
    ${\small \text{Bsl}_{\thinspace\text{enc.}}}$ & 0.098 & 11.18 & 0.272 & 0.378 / 0.019  \\
    ${\small \text{Bsl}_{\thinspace\text{0s}}}$   & 0.193 & 21.52 & 0.274 & 0.656 / 0.034 \\
    ${\small \text{Bsl}_{\thinspace\text{rnd.}}}$   & 0.238 & 26.77 & 0.277 & 0.715 / 0.037\\
    \midrule    
    ${\small \text{CHiVE}_{\thinspace\text{enc.}}}$  & 0.077 & 8.478 & 0.287 & 0.356 / 0.019 \\
    ${\small \text{CHiVE}_{\thinspace\text{0s}}}$ & 0.173 & 19.16 & 0.294 & 0.515 / 0.027 \\
    ${\small \text{CHiVE}_{\thinspace\text{rnd.}}}$ &  0.214 & 24.37 & 0.298 & 0.585 / 0.031 \\
    \bottomrule
  \end{tabular}
  \label{tab:objective_results}
\end{table}
We also see that the absolute $F_0$ loss both when using the zero embedding and random embedding is substantially higher than when the encoded embedding is used. This is expected as these represent renditions of the utterance that are averaged or random, and as such may end up being quite different to the held-out example.\newline
The ratio between the $F_0$ and $c_0$ in the loss was not tuned, which might explain why the distinctive trend in the $F_0$ seems to be reversed for the $c_0$ terms, albeit less pronounced.   
Interestingly, the loss when zero embeddings are used is lower than when random embeddings are used.
This is in line with the intuition that zero embeddings represent an average, overall suitable, prosody, rather than an arbitrary point in the prosodic space.

\subsection{Prosody transfer between sentences}
\label{sec:prosody_transfer}

During training the encoder/decoder portions of the CHiVE model encode/decode the same text and speech.
At inference time, however, it is possible to encode one utterance to a sentence prosody embedding and then use that embedding to condition a decoder that generates prosodic features for an entirely different sentence. When CHiVE is run in this fashion, the sentence prosody embedding -- displayed in dark yellow at the bottom in Figure~\ref{fig:chive_decoder} -- represents the prosody of a reference sentence that was input to the encoder, while the sentence-, word-, syllable- and phone-level features -- displayed in yellow, green, pink and purple respectively -- represent another, new sentence, for the decoder to predict the prosodic features for. As a result, the prosody of the new sentence is guided by the prosody of the reference sentence. 

To illustrate the prosody transfer capabilities of CHiVE, we show $\log F_0$ curves for two pairs of sentences.
We use jokes in our examples, as they have a particular prosody pattern, clearly visible in the figures.
Note, though, that this is done for illustrative purposes only, and that prosody transfer is not restricted in any way to a particular type of sentence.

\begin{figure}[t]
  \centering
  \includegraphics[width=.46\textwidth]{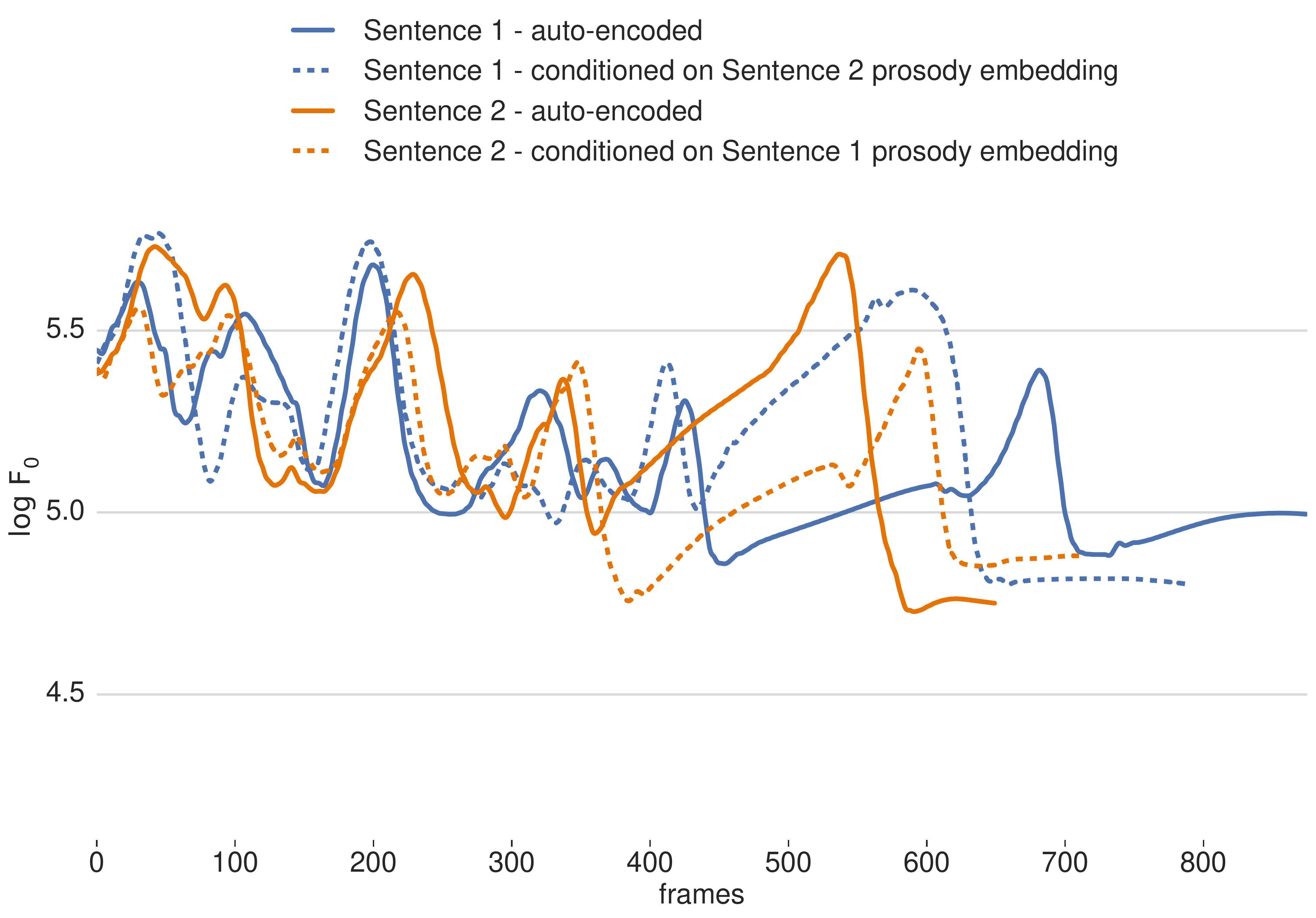}
  \caption{
    The $\log F_0$ curves for Sentence 1 \textit{``What do you call a boomerang that doesn't come back? A stick!''} and Sentence 2 \textit{``What's orange and sounds like a parrot? A carrot.''} conditioned on their own sentence prosody embedding (auto-encoded) and each other's sentence prosody embedding.}
  \label{fig:transfer_boomerang_parrot}
\end{figure}

\begin{figure}[t]
  \centering
  \includegraphics[width=.46\textwidth]{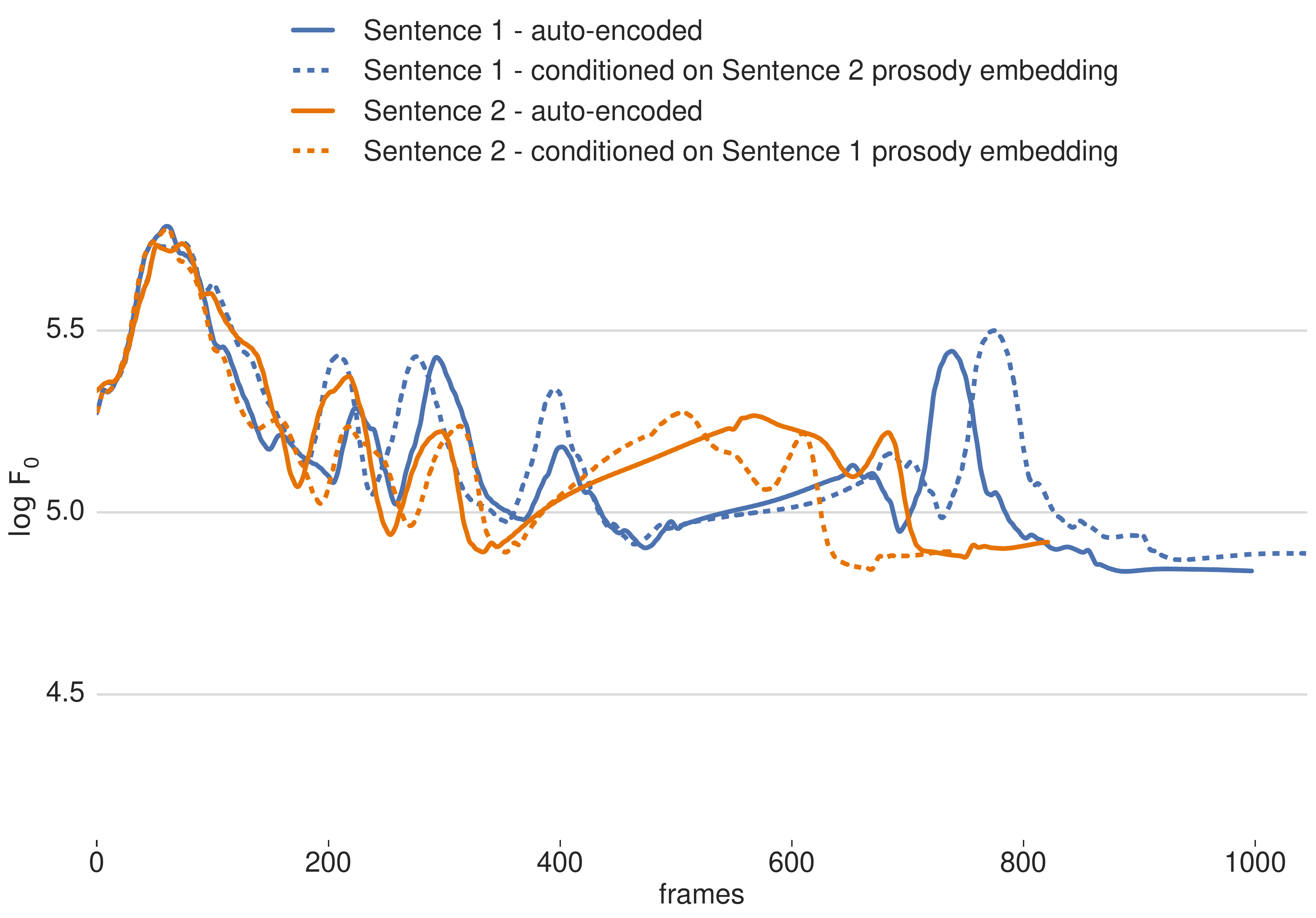}
  \caption{The $\log F_0$ curves for Sentence 1 \textit{``What do you get when you put a vest on an alligator? An investigator.''} and Sentence 2 \textit{``What do you get from a pampered cow? Spoiled milk.''} conditioned on their own sentence prosody embedding (auto-encoded) and each other's sentence prosody embedding.}
  \label{fig:transfer_alligator_cow}
\end{figure}

Figures~\ref{fig:transfer_boomerang_parrot}~and~\ref{fig:transfer_alligator_cow} both show the $\log F_0$ curves of two sentences.
Two curves are shown for each sentence: the solid lines are the result of running CHiVE in the auto-encoder setting; the dashed lines are the result of conditioning the decoder for one sentence on the sentence prosody embedding of the other.\newline
If no prosody transfer would take place, the pair of curves per sentence -- the two blue lines, and the two orange lines in Figures~\ref{fig:transfer_boomerang_parrot}~and~\ref{fig:transfer_alligator_cow} -- would overlap.
However, as can be observed from the figures, the opposite happens, where the dashed lines follow the solid lines of different colour.\newline
In Figure~\ref{fig:transfer_boomerang_parrot}, the dashed orange line, instead of closely following the solid orange line, is guided by the solid blue line as well.
It clearly mimmicks the solid blue line near the end, where the punch line is delivered.
The two curves do not perfectly overlap as there is a mismatch in terms of lengths (in words and syllables), which indicates that, even if the two sentences do not perfectly align in terms of syllable structure, meaningful prosody transfer can take place.\newline
In Figure~\ref{fig:transfer_alligator_cow} we see a similar effect, where at the start, both the solid blue/dashed orange curves and the solid orange/dashed blue curves are closely tied.
The transfer is not perfect, as can be seen, e.g., near the end, where the two blue line ends up much like each other, without, apparently, following the guidance of the reference prosody embedding, with something similar happening to the orange curves.

As these examples illustrate, the variational CHiVE model can be used to transfer prosody from one sentence to another.
Additional research is needed for a more rigorous analysis, to see what aspects affect succes or failure of transfer, and if transfer can be extended more generally to, e.g., speaking style.
Initial investigations with texts containing the same number of words suggests that specific prosodic patterns transfer between utterances, even when the number of syllables per word is different, and that prominence transfers to the correct syllable in the target sentence. 

%!TEX root = chive.tex
\section{Conclusion}
\label{sec:conclusions}

%% What did we do
Our model, CHiVE, a linguistically driven dynamic hierarchical conditional variational auto-encoder model, meets the objective of yielding a prosodic space from which meaningful prosodic features can be sampled to generate a variety of valid prosodic contours.
%% How do we perform
We showed that the prosody generated by the model, when using the mean of the distribution modeled in the variational layer (the zero embedding) performs as well as current state of the art. We also showed that the hierarchical structure yields speech with a higher variety in natural pitch contours for a given text when compared to a model without this hierarchy.
Lastly, we illustrated that the sentence embedding produced by the encoder, that captures the prosody of one utterance, can be used to transfer the prosody to another text.\newline
%% Larger picture. What do we learn?
CHiVE can capture natural prosodic variations originally not inferable from the linguistic features it gets as input. This opens opportunities for speech synthesis systems to sound more natural, particularly when synthesizing longer pieces of text -- e.g., when a digital assistant reads out a long answer or an entire audiobook -- as in these settings repeated use of a default prosody can become tiring to listen to.\newline
As the space captured by our hierarchical variational model is continuous -- arbitrary points yield natural sounding prosody -- it would be natural to want to control this space, e.g., by manually selecting a point in it rather than random sampling.
Future work will concentrate on making the utterance embedding space interpretable in a meaningful way.

%% Limitations
% Finally, even though we show transfer of prosody between sentences to be possible, it is limited to sentences that are comparable in terms of syntactic structure and number syllables.
% Further research is needed to extend this to a broader notion, such as perhaps \emph{intent}, where the prosody of a target sentence might be guided by the intent of a reference sentence, as deduced from its prosody. 

%% Future work

% tomkenter: I am not sure what this next sentence is about...
% Future work will examine the finer aspects of controllability that can be achieved by manipulating the syllable level activations.

\newpage
%% %%%% Acknowledgements shoud be commented out for double blind review process.
\section*{Acknowledgments}
We would like to acknowledge contributions from the Google AI wider TTS research community and DeepMind, calling out specifically contributions from DeepMind in building \(c_0\) conditioned WaveNet Models, and helpful discussion and technical help with
evaluation from the Sound Understanding team.

\bibliography{chive}
\bibliographystyle{icml2019}

\end{document}